%% file: main.tex
\documentclass{article}

\usepackage[final,main]{neurips_2026}

\usepackage[utf8]{inputenc}
\usepackage[T1]{fontenc}
\usepackage{xcolor}
\definecolor{ourHi}{RGB}{222,235,250}
\definecolor{oracleHi}{RGB}{242,242,242}
\definecolor{linkblue}{RGB}{30,90,200}
\usepackage[colorlinks=true,linkcolor=linkblue,citecolor=linkblue,urlcolor=linkblue]{hyperref}
\usepackage{url}
\usepackage{booktabs}
\usepackage{capt-of}
\usepackage{multirow}
\usepackage{colortbl}
\usepackage{amsmath}
\usepackage{amssymb}
\usepackage{mathtools}
\usepackage{amsfonts}
\usepackage{bbm}
\usepackage{nicefrac}
\usepackage{microtype}
\usepackage{pifont}
\usepackage{graphicx}
\usepackage{algorithm}
\usepackage{algpseudocode}
\usepackage{wrapfig}
\usepackage{enumitem}
\newcommand{\cmark}{\textcolor{green!60!black}{\ding{51}}}
\newcommand{\xmark}{\textcolor{red!75!black}{\ding{55}}}

\title{SARL: Optimizing the Path of Reasoning with Structure Aware Reinforcement Learning}

\author{
\mdseries Yifan Wang$^{1}$ \quad Bolian Li$^{1}$ \quad David Cho$^{1}$ \\
\mdseries Ruqi Zhang$^{1}$ \quad Fanping Sui$^{2}$ \quad Ananth Grama$^{1}$ \\
\\
$^{1}$Purdue University \quad
$^{2}$Texas Instruments
}

\begin{document}

\maketitle

\begin{abstract}
Reinforcement learning is critical to improving large reasoning models, but its success relies heavily on verifiable rewards (RLVR), making it hard to use in open-ended domains where correctness is ambiguous and cannot be verified. Moreover, reasoning trajectories remain largely unconstrained, and optimizing solely towards the final answer can favor early exploitation over generalization. In this work, we investigate whether general reasoning ability can be improved by teaching models \textbf{\textit{how to think}} (the structure of reasoning) rather than \textbf{\textit{what to produce}} (the outcome of reasoning), and we extend traditional RLVR to open-ended settings. We introduce \textbf{S}tructure \textbf{A}ware \textbf{R}einforcement \textbf{L}earning (\textbf{SARL}), a label-free framework that constructs per-response reasoning maps from intermediate thinking steps and rewards their reasoning topology. SARL shifts supervision from destination to path, encouraging reasoning trajectories that are both locally coherent and globally efficient. On verifiable math tasks, SARL outperforms prior label-free RL baselines and even exceeds RL methods with ground truth supervision, with average gains of +9.1\% under PPO and +11.6\% under GRPO across four math benchmarks, with particularly large improvements on AIME25 (+35.5\% with PPO and +44.7\% with GRPO). On non-verifiable open-ended tasks, SARL achieves average gains of +34.6\% under PPO and +30.4\% under GRPO on WildBench across five task categories, outperforming prior label-free RL methods and DPO, which relies on additional preference labels. Beyond strong performance, SARL exhibits substantially lower KL divergence and higher policy entropy, indicating more stable and exploratory training dynamics. Code and data are available at \href{https://github.com/cacayaya/SARL}{\textcolor{linkblue}{\texttt{github.com/cacayaya/SARL}}}.
\end{abstract}

\input{sections/introduction}
\input{sections/related-work}

\input{sections/method}
\input{sections/experiment}
\input{sections/limitation}
\input{sections/conclusion}

\bibliography{main}
\bibliographystyle{plainnat}

\appendix
\input{sections/appendix}

\end{document}

%% file: sections/introduction.tex
\section{Introduction}

Recently, Large Reasoning Models (LRMs) have demonstrated transformative capabilities in tackling complex, reasoning-intensive tasks, especially in mathematics and code generation~\citep{guo2025deepseek,yang2025qwen3}. Central to this progress is the Chain-of-Thought (CoT) paradigm, where models generate explicit, step-by-step reasoning before producing a final answer~\citep{wei2022chain,wang2022self}. These reasoning capabilities are typically grounded in supervised fine-tuning (SFT) and, more crucially, in reinforcement learning (RL)~\citep{ouyang2022training,shao2024deepseekmath}. Through RL training, models develop stronger reasoning abilities and exhibit sophisticated ``thinking'' behaviors such as self-reflection and multi-step planning~\citep{lightman2023let,wang2024math}.

\begin{figure}[thp]
\centering
\includegraphics[width=\linewidth]{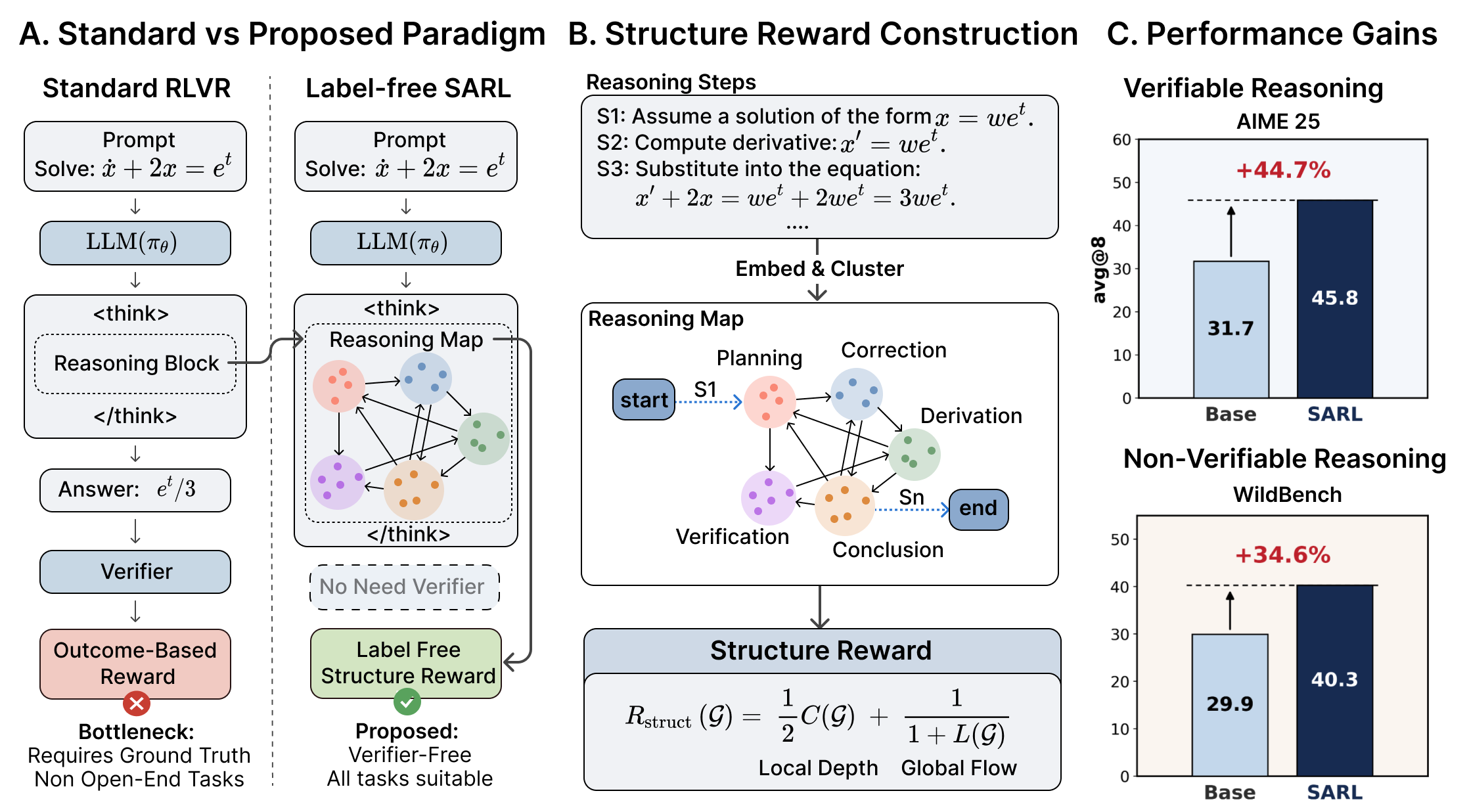}
\vspace{-15pt}
\caption{Overview of Structure Aware Reinforcement Learning (SARL). Left: SARL replaces outcome based supervision with a label free structure reward computed from the model's reasoning process. Middle: construction of the SARL Reasoning Map and Structure Reward. Right: Performance improvements on verifiable and open-ended tasks.}
\vspace{-5pt}
\label{fig:marl}
\end{figure}

However, the current success of RL training is largely tethered to the scalability of Reinforcement Learning from Verifiable Rewards (RLVR) framework~\citep{guo2025deepseek,shao2024deepseekmath}, which relies heavily on tasks with objective ground truths and automated verifiers to provide reward signals. This dependency creates a bottleneck of verifiability: in domains where ground truth is ambiguous, expensive to label, or non-existent, such as open-ended strategic planning or complex philosophical inquiry, reliable reward signals are largely unavailable.

Furthermore, even finer-grained approaches such as Process Reward Models (PRMs)~\citep{lightman2023let,ma2023let,khalifa2025process} remain fundamentally outcome-oriented: they assess how much each intermediate step contributes to reaching the correct final answer, rather than directly optimizing the structural organization of the reasoning process itself. As a result, the overall topology of reasoning trajectories remains unshaped. Moreover, PRMs are also prohibitively expensive to deploy in practice, as they require dense step-level annotations that are costly to collect and do not generalize to open-ended tasks without clear verification criteria. This raises a central question:

\begin{quote}
\textit{Can complex reasoning be optimized not through supervision derived from ground-truth answers, but by directly shaping how the model thinks?}
\end{quote}

A compelling answer to this question is suggested by recent empirical work on the reasoning structural properties of Large Reasoning Models. Reasoning graphs extracted from LRMs exhibit measurable \emph{small-world} topology~\citep{watts1998collective}, with dense local clustering coexisting with short global path lengths, and these structural features correlate positively with reasoning accuracy~\citep{minegishi2025topology,feng2025characterizes,tan2025shape}. From the perspective of complex network theory, such organization balances local specialization with global integration, allowing efficient communication across distinct functional substructures of a reasoning trace. Together, these findings suggest that small-world topology is not merely a byproduct of good reasoning but a structural property that actively underlies it.

In this work, we introduce Structure Aware Reinforcement Learning (SARL), a label-free framework that takes this structural insight as a direct training signal. Rather than relying on ground-truth answers or step-level annotations, SARL constructs per-response reasoning maps from intermediate thinking steps and optimizes its small-world topology as a reward, enabling RL training across both verifiable and open-ended domains.

Our main contributions are as follows:
\begin{itemize}
    \item We formulate the structural quality of reasoning trajectories as an intrinsic, label-free training signal, establishing a new optimization objective orthogonal to outcome-based rewards.
    \item We show that SARL is effective in both mathematical and open-ended reasoning, outperforming prior label-free baselines and even surpassing RL with ground-truth supervision.
    \item We observe that SARL yields lower KL divergence and higher policy entropy throughout training, suggesting more stable training dynamics and sustained exploration.
\end{itemize}

%% file: sections/related-work.tex
\section{Related Work}
\paragraph{Topology of Reasoning in Large Reasoning Models.}
A growing line of work characterizes reasoning in Large Reasoning Models (LRMs) through the structural organization of their reasoning trajectories.
The \emph{small-world} property~\citep{watts1998collective}, where dense local clustering coexists with short global path lengths, has emerged as a particularly informative way to characterize these trajectories.
\citet{minegishi2025topology} extract reasoning graphs from the hidden-state trajectories of LRMs and find that distilled reasoning models exhibit substantially higher cyclicity, diameter, and small-world index than base models, with these structural advantages scaling with task difficulty and correlating positively with accuracy.
From a geometric perspective, \citet{zhou2025geometry} model chain-of-thought reasoning as smooth flows in the model's representation space, showing that logical transitions induce structured geometric trajectories governed by the semantics of intermediate propositions.
At the level of trace analysis, \citet{feng2025characterizes} introduce graph-based structural metrics for CoT traces and demonstrate that the fraction of abandoned reasoning branches, rather than trace length or self-review frequency, is the strongest predictor of correctness, providing causal evidence that structural organization shapes reasoning outcomes.
\citet{tan2025shape} further show that topological data analysis features of reasoning traces carry substantially higher predictive power for reasoning quality than standard graph-connectivity metrics.
Collectively, these studies establish a robust empirical link between the structural organization of reasoning trajectories and reasoning quality.
Our work takes the natural next step: rather than analyzing these structural properties post hoc, we optimize the small-world topology of reasoning graphs, built from individual steps as a direct intrinsic reasoning quality signal.

\paragraph{Label Free RLVR}
Reinforcement Learning from Verifiable Rewards (RLVR) has emerged as the dominant post-training paradigm for LRMs~\citep{guo2025deepseek}, using verifiable outcome rewards in domains where correctness can be automatically evaluated, primarily in mathematics and code.
However, current RLVR-based training has two key limitations: (1) it relies on verifiable ground-truth rewards, making it inapplicable to open-ended domains such as creative writing, strategic planning, or advisory reasoning, where no automated verifier exists; and (2) it offers limited control over the structure of intermediate reasoning, as supervision operates solely at the level of correctness of final answer.Alternative label-free approaches have emerged to enable RLVR without ground-truth annotations.
TTRL~\citep{zuo2025ttrl} uses majority voting across rollouts to construct pseudo-labels as reward signals, which works for closed-form tasks but cannot handle open-ended settings where no consensus answer exists.
\citet{xin2025surrogate} propose surrogate rewards based on output format compliance and response length, though this remains restricted to math tasks with prescribed format templates. Meanwhile, EMPO~\citep{zhang2025right} and concurrent work~\citep{agarwal2025unreasonable} minimize the predictive entropy of model outputs, using the model's own uncertainty as a label-free reward signal. Similarly, Intuitor~\citep{zhao2025learning} utilizes self-certainty scores measured via KL divergence to a uniform distribution as reward, and formulates the Reinforcement Learning from Internal Feedback (RLIF) framework. Beyond these methods, our approach derives the label free reward signal from the structure of the model's reasoning processes, and is applicable to not only verifiable tasks but also to open-ended and complex tasks in broader and real-world reasoning settings. Table~\ref{tab:comparison} summarizes the key differences between these label-free approaches. 

\begin{table}[ht]
\centering
\caption{Comparison of \emph{label-free} RL methods. ``Algorithm-agnostic'' indicates whether the reward signal 
can be used with any RL algorithm (e.g., PPO) beyond group-based methods (e.g., GRPO).}
\label{tab:comparison}
\renewcommand{\arraystretch}{1.1}
\setlength{\tabcolsep}{3pt}
\small
\begin{tabular}{lcccc}
\toprule
\textbf{Method}
  & \textbf{Reward Signal}
  & \textbf{Algorithm-agnostic}
  & \textbf{Open-ended}
  & \textbf{Verifiable} \\
\midrule
EMPO~\citep{zhang2025right}
  & Entropy minimization
  & \xmark
  & \cmark & \cmark \\
TTRL~\citep{zuo2025ttrl}
  & Majority voting
  & \xmark
  & \xmark & \cmark \\
Format \& Length~\citep{xin2025surrogate}
  & Format + length
  & \cmark
  & \xmark & \cmark \\
\midrule
\textbf{SARL (Ours)}
  & \textbf{Reasoning Structure}
  & \textbf{\cmark}
  & \textbf{\cmark} & \textbf{\cmark} \\
\bottomrule
\end{tabular}
\end{table}

%% file: sections/method.tex
\section{Methodology}
\label{sec:method}

In this section, we introduce Structure Aware Reinforcement Learning (SARL), a label free framework that rewards the topology of intermediate reasoning. We begin with the problem formulation, transitioning from outcome based RL to a structural training objective (Section~\ref{subsec:problem}). We then describe the construction of the reasoning map from thinking steps and the transitions of latent reasoning functions (Section~\ref{subsec:mapcon}). Finally, we define the structure reward (Section~\ref{subsec:reward}). The complete pipeline is summarized in Algorithm~\ref{alg:sarl}.

\begin{algorithm}[htp]
\caption{Structure Aware Reinforcement Learning}
\label{alg:sarl}
\begin{algorithmic}[1]
\Require Policy $\pi_\theta$, embedding model $\mathcal{M}$, dataset $\mathcal{D}$, RL algorithm $\mathcal{A}$, rollouts per question $N$
\For{each training iteration}
    \State Sample batch $\{x_i\}$ from $\mathcal{D}$
    \For{each question $x_i$}
        \State Generate $N$ trajectories $\{\tau_i^{(n)}\}_{n=1}^{N} \sim \pi_\theta(\cdot \mid x_i)$
        \For{each trajectory $\tau_i^{(n)}$}
            \State Extract steps $\{s_t\}$ from \texttt{<think>} block; embed: $\mathbf{e}_t \leftarrow \mathcal{M}(s_t)/\|\mathcal{M}(s_t)\|$
            \State Cluster $\{\mathbf{e}_t\}$ into $K$ nodes $\mathcal{V}=\{v_1,\ldots,v_K\}$; assign each $s_t$ to $v(s_t)\in\mathcal{V}$
            \State Build $\mathcal{G}(\tau_i^{(n)})$: add edge $(v(s_t),v(s_{t+1}))$ for each $v(s_t)\neq v(s_{t+1})$
            \State Compute structure reward: $r_i^{(n)} \leftarrow \mathrm{SR}\!\left(\mathcal{G}(\tau_i^{(n)})\right)$
        \EndFor
    \EndFor
    \State Update $\theta \leftarrow \mathcal{A}\!\left(\theta,\{(\tau_i^{(n)}, r_i^{(n)})\}\right)$
\EndFor
\end{algorithmic}
\end{algorithm}

\subsection{Problem Formulation}
\label{subsec:problem}

Let $x$ denote an input question and let $\pi_\theta$ be a language model policy parameterized by $\theta$. Following the chain-of-thought paradigm, the model produces a trajectory
\[
\tau = (s_1, s_2, \ldots, s_T, a),
\]
where $s_1, \ldots, s_T$ are intermediate reasoning steps inside the \texttt{<think>} block and $a$ is the final answer. Standard RLVR optimizes an outcome level objective of the form
\begin{equation}
    \max_\theta \;\mathbb{E}_{x \sim \mathcal{D},\;\tau \sim \pi_\theta(\cdot \mid x)}
    \bigl[\,r(a, x)\,\bigr],
    \label{eq:rlvr}
\end{equation}
where $r(a, x)$ is a verifiable reward that evaluates the final answer $a$ against the question $x$, e.g., via exact match, execution, or format checking. While effective in domains where such verification is available, Eq.~\eqref{eq:rlvr} conditions the reward entirely on the final answer and provides no direct signal about the quality of the reasoning process itself.

SARL replaces the outcome reward with a structural reward computed from the generated reasoning trajectory. For each rollout, we build a reasoning map $\mathcal{G}(\tau) = (\mathcal{V}, \mathcal{E})$ whose nodes represent latent reasoning functions and whose edges represent transitions between them. The SARL objective is
\begin{equation}
    \max_\theta \;\mathbb{E}_{x \sim \mathcal{D},\;\tau \sim \pi_\theta(\cdot \mid x)}
    \bigl[\,\mathrm{SR}\!\bigl(\mathcal{G}(\tau)\bigr)\,\bigr],
    \label{eq:objective}
\end{equation}
where $\mathrm{SR}$ denotes the structure reward and measures the small-world organization of the resulting graph. Crucially, Eq.~\eqref{eq:objective} does not rely on any verifier $r(a, x)$, allowing the policy to be trained even when correctness labels are unavailable or expensive to obtain.

\subsection{Reasoning Map Construction}
\label{subsec:mapcon}
\paragraph{Step Extraction and Embedding.}
We first extract the content of the <think> block and segment it into individual reasoning steps, where each step corresponds to a coherent sentence naturally delimited by the model's output formatting (e.g., \texttt{\textbackslash n\textbackslash n} for Qwen3-4B). Then, each step $s_t$ is mapped to a unit normalized embedding $\mathbf{e}_t \in \mathbb{R}^d$,
\[
    \mathbf{e}_t = \frac{\mathcal{M}(s_t)}{\|\mathcal{M}(s_t)\|_2},
\]
where $\mathcal{M}$ is a separate text embedding model.\footnote{We use \texttt{Qwen/Qwen3-Embedding-0.6B} for all experiments.}

\paragraph{Latent reasoning functions.}
A reasoning trajectory typically consists of multiple reasoning functions, such as planning, derivation, calculation, verification, correction, and others. Importantly, the same reasoning function may appear repeatedly at different positions within a single trajectory. For example, multiple calculation steps may occur throughout the reasoning process, each operating on different intermediate quantities while serving the same functional role. SARL clusters the step embeddings $\{\mathbf{e}_t\}_{t=1}^T$ into $K$ latent reasoning functions,
\begin{equation}
    \mathcal{V} = \{v_1, v_2, \ldots, v_K\}.
\end{equation}
Each node $v_k$ represents one latent reasoning function shared by steps
that perform similar operations, and we use
$v(s_t) \in \mathcal{V}$ to denote the node assigned to step $s_t$.
Clustering is performed independently for each response.
We consider both KMeans and HDBSCAN for this step; the specific settings
and the regimes in which each choice is more suitable are discussed in
Appendix~\ref{app:hdbscan}.

\paragraph{Reasoning transitions.}
The edge set $\mathcal{E}$ captures transitions between reasoning functions. We add an undirected edge $(v(s_t), v(s_{t+1})) \in \mathcal{E}$ whenever two consecutive steps map to different nodes, i.e., $v(s_t) \neq v(s_{t+1})$. Repeated transitions between the same pair of reasoning functions contribute only one edge. We use an undirected graph because the reward is intended to measure structural organization and connectivity rather than directional control flow.

\subsection{Structure Reward}
\label{subsec:reward}
\paragraph{Small-world prior.}
SARL is motivated by the observation that effective reasoning should balance local specialization with global integration. This suggests two desirable properties: high local clustering, which reflects coherent functional substructures, and short global path lengths, which reflect efficient transitions across different reasoning functions.

Let $\mathcal{N}(v_k)$ denote the neighbor set of node $v_k$, and let $\mathcal{V}_{\geq 2} = \{v_k \in \mathcal{V} : |\mathcal{N}(v_k)| \geq 2\}$ be the set of nontrivial nodes. We define the average clustering coefficient as
\begin{equation}
    C(\mathcal{G}) \;=\; \frac{1}{|\mathcal{V}_{\geq 2}|}
    \sum_{v_k \in \mathcal{V}_{\geq 2}}
    \frac{\bigl|\{(v_i,v_j)\in\mathcal{E} : v_i,v_j\in\mathcal{N}(v_k)\}\bigr|}
         {|\mathcal{N}(v_k)|\,\bigl(|\mathcal{N}(v_k)|-1\bigr)/2},
    \label{eq:clustering}
\end{equation}
which measures how strongly neighboring reasoning functions form locally coherent modules. We also define the average shortest path length over all reachable node pairs $\mathcal{P} = \{(v_i,v_j) : v_i \neq v_j,\; v_j \text{ reachable from } v_i\}$ as
\begin{equation}
    L(\mathcal{G}) \;=\; \frac{1}{|\mathcal{P}|} \sum_{(v_i,v_j)\in\mathcal{P}} \delta(v_i,v_j),
    \label{eq:pathlength}
\end{equation}
where $\delta(v_i,v_j)$ denotes hop count distance. Lower values of $L(\mathcal{G})$ indicate more efficient communication across distinct reasoning functions.
    
The classical small-world index $\sigma = (C(\mathcal{G})/C_{\mathrm{rand}}) / (L(\mathcal{G})/L_{\mathrm{rand}})$~\citep{watts1998collective}, even under the analytical approximation $C_{\mathrm{rand}}\approx\bar{k}/n$ and $L_{\mathrm{rand}}\approx\ln n/\ln\bar{k}$~\citep{humphries2008network}, becomes numerically unstable when $\bar{k}\approx 1$, precisely the sparse regime of the small reasoning graphs that arise in practice. We therefore define the structure reward (SR) as the equally weighted combination of the two classical small-world quantities, the clustering coefficient $C(\mathcal{G})$ and the average shortest path length $L(\mathcal{G})$, each mapped to $[0,\tfrac{1}{2}]$ so that they enter the reward on a common scale and the total reward is bounded in $[0,1]$:
\begin{equation}
    \mathrm{SR}(\mathcal{G}) \;=\;
    \frac{1}{2}\,C(\mathcal{G}) \;+\; \frac{1}{1 + L(\mathcal{G})}.
    \label{eq:reward}
\end{equation}
As shown in Eq.~\eqref{eq:clustering} and Eq.~\eqref{eq:pathlength}, $C(\mathcal{G})$ captures local specialization, while $L(\mathcal{G})$ captures global efficiency. Together, the two terms discourage traces that collapse into a single reasoning function or drift through long unstructured chains of transitions.

    

%% file: sections/experiment.tex
\section{Experiments}
\label{sec:experiments}

In this section, we present empirical results to demonstrate the effectiveness of SARL. We first describe the setup (Section~\ref{subsec:setup}), then present results on verifiable mathematical reasoning (Section~\ref{subsec:math_results}) and open-ended reasoning (Section~\ref{subsec:openrubrics_results}), and finally provide analysis and ablation studies (Section~\ref{subsec:analysis}).

\subsection{Setup}
\label{subsec:setup}

\paragraph{Model and Datasets.}
We conduct all experiments using Qwen3-4B~\citep{yang2025qwen3}, a strong reasoning model that generates explicit reasoning trajectories within \texttt{<think>} blocks. We evaluate SARL under two distinct training regimes that cover both \emph{verifiable} and \emph{non-verifiable} reasoning domains. Dataset filtering and preparation details are provided in Appendix~\ref{app:datasets}.

\begin{itemize}[leftmargin=2em,itemsep=2pt,topsep=2pt]
    \item \textbf{Verifiable reasoning (Math).}
    We train on mathematical reasoning problems drawn from historical AIME competitions (1983--2024). Unlike simpler math tasks such as GSM8K~\citep{cobbe2021training} or MATH500~\citep{hendrycks2021measuring}, AIME problems demand long, complex reasoning, producing trajectories whose topological structure is rich enough to be meaningful, making them particularly well-suited for SARL training.

    \item \textbf{Non-verifiable reasoning (Open-ended).}
    We train on OpenRubrics-v2~\citep{liu2025openrubrics}, a large-scale preference dataset spanning diverse domains including creative writing, planning, coding, advice, and analytical reasoning. Unlike mathematical tasks, these problems lack explicit correctness labels.
\end{itemize}

\paragraph{Baselines.}
For the verifiable setting (Math Reasoning), we compare against two prominent label-free RL baselines: EMPO~\citep{zhang2025right} and TTRL~\citep{zuo2025ttrl}. For the non-verifiable setting (Open-Ended Reasoning), we additionally compare against DPO~\citep{rafailov2023direct} training on preference data. TTRL is not applicable to open-ended reasoning, as it relies on majority voting that assumes a consensus which may not exist in such settings.

\paragraph{Training and Evaluation.}
All reinforcement learning experiments are implemented using the veRL framework~\citep{sheng2024hybridflow}. Preference-based training with DPO is implemented using LlamaFactory~\citep{zheng2024llamafactory}. Full training details, including hyperparameters and optimization settings, are provided in Appendix~\ref{app:training}.

For evaluation, we assess mathematical reasoning performance on four benchmarks: MATH500~\citep{hendrycks2021measuring}, AIME25~\citep{balunovic_srimatharena_2025}, AMC23~\citep{knoveleng_amc23} and Minerva Math~\citep{lewkowycz2022solving}. We adopt avg@$8$ as our primary evaluation metric, which provides a more stable assessment of model capability compared to single-run pass@1 scores. For open-ended reasoning evaluation, we employ the WildBench Elo rating system and task-category macro scores from the WildBench leaderboard~\citep{lin2024wildbench}.


\subsection{Verifiable Reasoning (Mathematical Tasks) Results}
\label{subsec:math_results}

We first validate the effectiveness of SARL in the well-established verifiable setting of mathematical reasoning, where RLVR methods are most commonly applied.

\begin{table*}[htp]
\centering
\small
\setlength{\tabcolsep}{6pt}
\renewcommand{\arraystretch}{1.15}
\begin{tabular}{lccccc}
\toprule
\textbf{Method} & \textbf{AIME25} & \textbf{AMC23} & \textbf{MATH500} & \textbf{Minerva} & \textbf{Avg ($\Delta$\%)} \\
\midrule
Base & 31.67 & 82.81 & 90.10 & 53.45 & 64.51 \\
\midrule
\multicolumn{6}{l}{\textit{PPO}} \\
\quad w/ Ground-Truth$^{\dagger}$        & 41.67 & \underline{86.56} & 92.75 & 59.74 & 70.18~(+8.8\%) \\
\rowcolor{ourHi}
\quad w/ SARL~(\textbf{Ours})            & 42.92 & 85.00 & 92.53 & 61.08 & 70.38~(+9.1\%) \\
\midrule
\multicolumn{6}{l}{\textit{GRPO}} \\
\quad w/ Ground-Truth$^{\dagger}$        & \textbf{46.67} & 84.38 & 93.15 & \textbf{62.45} & \underline{71.66~(+11.1\%)} \\
\quad w/ EMPO                            & 44.58 & \underline{86.56} & \underline{93.23} & 61.44 & 71.45~(+10.8\%) \\
\quad w/ TTRL                            & 42.91 & \underline{86.56} & 93.15 & 61.86 & 71.12~(+10.2\%) \\
\rowcolor{ourHi}
\quad w/ SARL~(\textbf{Ours})            & \underline{45.83} & \textbf{87.50} & \textbf{93.30} & \underline{61.99} & \textbf{72.16~(+11.9\%)} \\
\bottomrule
\end{tabular}
\caption{Results on mathematical reasoning benchmarks. \textbf{Bold} marks the best score in each column and \underline{underline} marks the second best. $^{\dagger}$ indicates methods that train with ground-truth correctness labels (oracle); all other methods are fully label-free. SARL achieves the best overall average under both algorithms and matches or surpasses the ground-truth supervision.}
\label{tab:math_main}
\end{table*}
\vspace{-10pt}

\paragraph{Structure reward closes the gap to ground-truth supervision.}
Table~\ref{tab:math_main} shows that SARL matches or surpasses ground-truth RL under both PPO and GRPO, without ever observing a correctness label. Under GRPO, SARL achieves the best performance among all label-free RL methods, reaching an average relative improvement of +11.9\%, outperforming EMPO (+10.8\%) and TTRL (+10.2\%), and even exceeding ground-truth RL (+11.1\%). Under PPO, SARL achieves a relative improvement of +9.1\%, again surpassing ground-truth PPO training (+8.8\%). These results indicate that structural signals alone can provide sufficiently strong learning guidance to rival, and in some cases exceed, correctness-based supervision. Notably, unlike EMPO and TTRL, which rely on group-level optimization and are restricted to GRPO-style training, SARL generalizes across both PPO and GRPO frameworks. This algorithm-agnostic compatibility highlights its practical applicability across diverse reinforcement learning pipelines.

The largest gains from SARL training occur on \textbf{AIME25}, the most challenging benchmark requiring long multi-step mathematical derivations. Under GRPO, performance improves from 31.67 to \textbf{45.83}, corresponding to a gain of +14.16 points (+45\% relative improvement). Such improvements are substantially larger than those observed on easier benchmarks, like MATH500 and AMC23, suggesting that structural constraints become increasingly beneficial as reasoning becomes more complex. 

\subsection{Non-Verifiable Reasoning (Open-Ended Tasks) Results}
\label{subsec:openrubrics_results}
A core motivation of our work is to enable reinforcement learning beyond verifiable domains using label-free structural rewards. To test this, we apply SARL on OpenRubrics-v2~\citep{liu2025openrubrics}, a diverse open-ended QA dataset.

\begin{table*}[htp]
\centering
\small
\setlength{\tabcolsep}{6pt}
\renewcommand{\arraystretch}{1.15}
\begin{tabular}{lcccccc}
\toprule
\textbf{Method} & \textbf{Creative} & \textbf{Planning} & \textbf{Math} & \textbf{Info} & \textbf{Code} & \textbf{WB Score ($\Delta$\%)} \\
\midrule
Base & 51.01 & 36.23 & 16.35 & 48.71 & 14.72 & 29.91 \\
\midrule
\multicolumn{7}{l}{\textit{DPO}} \\
\quad w/ preference labels$^{\dagger}$ & 51.16 & 37.82 & 17.06 & 48.37 & 14.34 & 30.34~(+1.4\%) \\
\midrule
\multicolumn{7}{l}{\textit{PPO}} \\
\rowcolor{ourHi}
\quad w/ SARL~(\textbf{Ours})          & \textbf{57.05} & \textbf{45.87} & \textbf{27.70} & \textbf{53.47} & \textbf{30.05} & \textbf{40.26~(+34.6\%)} \\
\midrule
\multicolumn{7}{l}{\textit{GRPO}} \\
\quad w/ EMPO                          & 51.01 & 36.11 & 17.62 & 45.69 & 12.92 & 29.20~(-2.4\%) \\
\rowcolor{ourHi}
\quad w/ SARL~(\textbf{Ours})          & \underline{55.34} & \underline{43.56} & \underline{26.83} & \underline{51.98} & \underline{29.91} & \underline{39.01~(+30.4\%)} \\
\bottomrule
\end{tabular}
\caption{WildBench evaluation results for open-ended reasoning. Methods are grouped by training algorithm (DPO, PPO, GRPO). \textbf{Bold} marks the best score in each column and \underline{underline} marks the second best. $^{\dagger}$ indicates methods that train with explicit human supervision (preference labels for DPO); SARL and EMPO are fully label-free. TTRL is omitted because it requires binary correctness labels and does not generalize to open-ended tasks.}
\label{tab:wildbench_main}
\end{table*}
\vspace{-5pt}

\paragraph{Structure reward enables reinforcement learning beyond verifiable domains.}
SARL achieves substantial performance gains on open-ended tasks without any labels or preference signals: under PPO, the WB Score improves from 29.91 to 40.26 (+34.6\% relative improvement), and under GRPO to 39.01 (+30.4\%). Note that TTRL requires binary correctness labels and format-based rewards~\citep{xin2025surrogate} are math-specific, so neither generalizes here; we therefore compare against EMPO~\citep{zhang2025right} as the label-free baseline and DPO~\citep{rafailov2023direct} as a reference with preference supervision. Even with this extra supervision, DPO yields only a marginal improvement (+1.4\%), while EMPO slightly decreases performance (-2.4\%). In contrast, SARL consistently delivers large gains across all five task categories.

As shown in Table~\ref{tab:wildbench_main}, Planning improves by +26.6\% (PPO) and +20.2\% (GRPO), Creative writing by +11.8\% and +8.5\%, and Info/Advice by +9.8\% and +6.7\%. These categories have no single correct answer and rely on flexible judgment and knowledge synthesis, yet the structural prior still steers the model toward more coherent reasoning trajectories. The consistency of these gains suggests that small-world topology captures a fundamental quality of \emph{how to think}, rather than serving as a proxy for correctness, and thus generalizes well beyond the verifiable domain.

\subsection{Analysis and Ablations}
\label{subsec:analysis}

\paragraph{How SARL reshapes the reasoning map.}

\begin{figure}[htp]
\centering
\includegraphics[width=\linewidth]{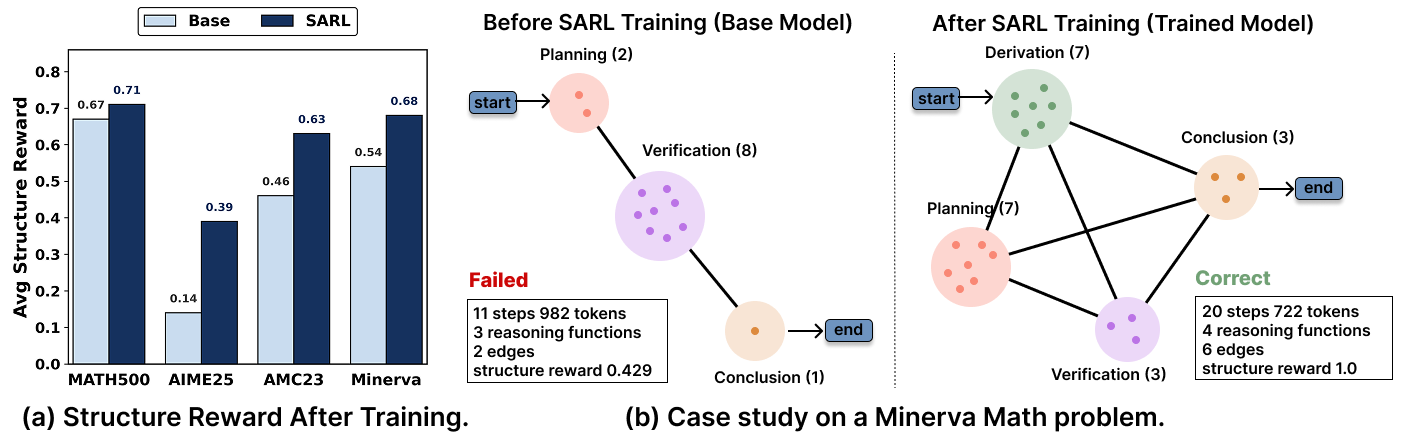}
\vspace{-12pt}
\caption{(a) Structure reward change across all benchmarks after SARL training. (b) Case study on a Minerva Math problem: reasoning topology before and after SARL training.}
\label{fig:structure_analysis}
\end{figure}
\vspace{-5pt}

Figure~\ref{fig:structure_analysis} analyzes how SARL reshapes reasoning structure on math benchmarks. Structure reward increases consistently across all four benchmarks after training, with the largest gain on AIME25 (+179\%), where the base model severely under-reasons (${\sim}21$ steps) and SARL encourages substantially deeper deliberation. On MATH500, SR rises despite nearly unchanged step count, indicating that gains come from better \emph{organization} of existing steps rather than more steps. These structural improvements co-occur with the accuracy gains in Table~\ref{tab:math_main} while response length \emph{decreases} (Table~\ref{tab:length_main}), confirming that SARL steers the model toward more efficient while effective reasoning. Figure~\ref{fig:structure_analysis}(b) illustrates this concretely using a problem from Minerva Math: the base model produces a near-linear chain dominated by Verification (8 of 11 steps, SR${=}0.43$, incorrect), while the trained model distributes reasoning across four functions with 6 edges (SR${=}1.0$, correct) in \emph{fewer} tokens (722 vs.\ 982).

\paragraph{Training dynamics analysis.}

\begin{figure}[htp]
\centering
\includegraphics[width=\linewidth]{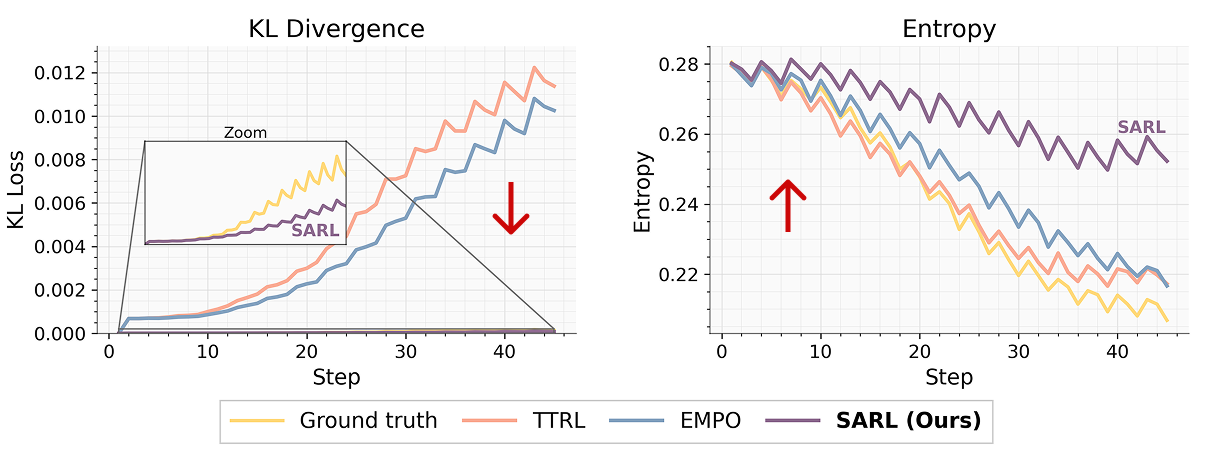}
\vspace{-15pt}
\caption{Training dynamics of different methods under GRPO.}
\label{fig:training_dynamics_main}
\end{figure}
\vspace{-5pt}

Figure~\ref{fig:training_dynamics_main} compares the GRPO training dynamics of SARL with both label-free baselines (TTRL, EMPO) and ground-truth supervision on the verifiable math task. SARL consistently maintains the smallest KL divergence to the reference policy throughout training, while the other methods exhibit a substantially larger policy drift. This behavior reflects a common failure mode of label-free RL methods: excessive policy drift can destabilize optimization and eventually lead to model collapse. 

At the same time, SARL maintains high policy entropy over most of training, indicating sustained exploration rather than early collapse to narrow reasoning patterns. In contrast, training directly on ground-truth rewards drives entropy drop quickly, suggesting premature exploitation that may hurt generalization. The central insight is that SARL improves reasoning through a stable optimization regime that combines minimal policy shift with relatively high exploration.

\paragraph{Length analysis.}
\begin{wraptable}{r}{0.45\linewidth}
\vspace{-20pt}
\centering
\small
\setlength{\tabcolsep}{3pt}
\renewcommand{\arraystretch}{1.12}
\begin{tabular}{lcc}
\toprule
\textbf{Method} & \textbf{Avg Tokens} & \textbf{$\Delta$ vs. Base} \\
\midrule
\multicolumn{3}{l}{\textbf{Verifiable reasoning (Math).}} \\
Base & 5736.6 & 0.0 \\
\rowcolor{ourHi}
PPO w/ SARL & 5075.6 & -661.0 \\
\rowcolor{ourHi}
GRPO w/ SARL & 4587.8 & -1148.8 \\
\midrule
\multicolumn{3}{l}{\textbf{Non-verifiable reasoning (Open-ended).}} \\
Base & 2677.91 & 0.00 \\
\rowcolor{ourHi}
PPO w/ SARL & 2303.17 & -374.74 \\
\rowcolor{ourHi}
GRPO w/ SARL & 2285.87 & -392.04 \\
\bottomrule
\end{tabular}
\caption{Response length comparison.}
\label{tab:length_main}
\vspace{-30pt}
\end{wraptable}

One possible concern is that the gains from SARL may come simply from generating longer responses. Table~\ref{tab:length_main} shows that this is not the case. SARL substantially reduces average response length in both open-ended and mathematical reasoning settings, while achieving much higher performance. These results suggest that the benefit comes from producing more efficient and better organized reasoning traces rather than simply extending length. Full length comparisons against other baselines are provided in Appendix~\ref{app:length_analysis}.

\begin{wraptable}{l}{0.38\linewidth}
\vspace{-10pt}
\centering
\small
\setlength{\tabcolsep}{3pt}
\renewcommand{\arraystretch}{1.12}
\begin{tabular}{lc}
\toprule
\textbf{Reward variant} & \textbf{Avg ($\Delta$\%)} \\
\midrule
Base                  & 64.51 \\
SARL w/ $L$ only      & 70.96~(+10.0\%) \\
SARL w/ $C$ only      & 71.97~(+11.6\%) \\
\rowcolor{ourHi}
SARL w/ $C + L$       & \textbf{72.16~(+11.9\%)} \\
\bottomrule
\end{tabular}
\caption{Ablation of the two structure reward terms ($C$: clustering coefficient; $L$: average shortest path length), averaged over MATH500, AIME25, AMC23, and Minerva.}
\label{tab:ablation_components}
\vspace{-15pt}
\end{wraptable}
\vspace{10pt}
\paragraph{Ablation: clustering coefficient vs. average shortest path length.}
The structure reward in Eq.~\eqref{eq:reward} combines two classical small-world quantities of $\mathcal{G}$: the clustering coefficient $C(\mathcal{G})$ and the average shortest path length $L(\mathcal{G})$. We retrain Qwen3-4B under GRPO on the same math setting using only one of $C$ or $L$ as the reward. As shown in Table~\ref{tab:ablation_components}, each component alone yields substantial gains over the base model, indicating that both capture strong, independently useful structural signals for reasoning, which aligns with prior observations on their role in reasoning~\citep{minegishi2025topology}. Combining them further improves performance, achieving the best average. Full results are provided in Appendix~\ref{app:ablation_components}.

%% file: sections/limitation.tex
\section{Limitations}
\label{sec:limitations}

SARL is most effective for tasks that involve sufficiently long and complex reasoning processes. For straightforward tasks with only a few steps, the benefit of enforcing structural organization may be limited. Although the small-world prior works on the benchmarks studied here, other domains may favor different structures. The reward is a proxy for reasoning quality and can in principle be exploited; the best-of-$n$ and graph-statistic checks in Appendix~\ref{app:analysis-reward-hacking} test particular failure modes but cannot establish the faithfulness or causal validity of individual traces. Performance also depends on step segmentation, embeddings, and clustering. Appendix~\ref{app:analysis-robustness} reports checks with another embedding model, clustering method, model scale, and model family. Open-ended results rely on automated judges, whose scores vary by evaluator; we report a second benchmark and multiple WildBench judges in Appendix~\ref{app:analysis-evaluation}. These limitations motivate more direct tests of reasoning faithfulness and broader robustness studies.

%% file: sections/conclusion.tex
\section{Conclusion}

In this work, we introduced structure aware reinforcement learning (SARL), a label free framework that improves reasoning by rewarding the organization of the reasoning process rather than the correctness of final answers. By constructing Reasoning Maps and optimizing their small world topology, SARL provides a scalable training signal that applies to both verifiable and non verifiable tasks. Empirically, SARL matches or surpasses ground truth based RL on mathematical reasoning benchmarks, outperforms existing label free baselines such as EMPO and TTRL, and delivers large gains on open ended WildBench evaluation without using correctness labels or preference supervision. Our analysis shows that SARL maintains minimal policy drift and consistently higher entropy throughout training, indicating a more stable optimization regime with sustained exploration. Taken together, these results suggest that reasoning structure itself is a strong and general learning signal, opening a path toward reinforcement learning beyond the bottleneck of verifiability.

%% file: sections/appendix.tex
\section{Implementation Details}
\label{sec:appendix}

\subsection{Clustering details for functional nodes}
\label{app:hdbscan}
We consider two practical choices for clustering step embeddings into latent reasoning functions: KMeans and HDBSCAN.

\paragraph{KMeans.}
KMeans is well suited to settings where the number of reasoning steps is not very large and the step structure is relatively regular. In our experiments, we use KMeans for the mathematical reasoning setting, where trajectories often follow a more regular progression such as setup, derivation, verification, and answer extraction, and we find KMeans empirically stable and reproducible across runs. Since KMeans requires specifying the number of clusters, we use $k \approx \sqrt{M}$, where $M$ is the number of reasoning steps in the current response.

\paragraph{HDBSCAN.}
HDBSCAN is preferable when the step structure is more heterogeneous because it can infer the number of clusters automatically and can better capture non spherical cluster geometry. In our experiments, we use HDBSCAN for the open ended setting, where the reasoning patterns are more diverse across prompts. For HDBSCAN, we use Euclidean distance, i.e., $d(\mathbf{e},\mathbf{e}')=\lVert \mathbf{e}-\mathbf{e}' \rVert_2$. We set $\text{min\_cluster\_size}=\max(2, \min(5, M/4))$ and $\text{min\_samples}=\text{min\_cluster\_size}-1$, where $M$ is the total number of step embeddings. Noise points (assigned label $-1$) are either grouped into a separate cluster or, if all points are noise, assigned to a default cluster.

\section{Experimental Setup}
\label{app:exp-setup}

\subsection{Data Preparation}
\label{app:datasets}

\paragraph{AIME Historical Problems (Math Setting).}
We use competition problems from the American Invitational Mathematics Examination (AIME) spanning 1983 to 2024, sourced from the publicly available \textit{AIME Problems 1983--2024} dataset on Kaggle.\footnote{\url{https://www.kaggle.com/datasets/tourist800/aime-problems-1983-to-2024}} This dataset contains 933 problems in total. Each problem is formatted as a single-turn prompt with the instruction ``Let's think step by step and output the final answer within \textbackslash boxed\{\}.'' appended to guide the model to solve the problem and box the final answer.

\paragraph{OpenRubrics-v2 (Open-Ended Setting).}
We use the \texttt{OpenRubrics/OpenRubric-v2} dataset from HuggingFace,\footnote{\url{https://huggingface.co/datasets/OpenRubrics/OpenRubric-v2}} which contains preference pairs across diverse task categories including creative writing, planning, coding, and advice.

We use only the \texttt{instruction} field as the training prompt and apply the following filtering pipeline to select high-quality prompts. First, we perform response length filtering: since this dataset contains reference responses, we use response length as a proxy to identify questions that elicit sufficiently complex reasoning. We require both reference responses should be longer than 512 tokens. Responses shorter than 512 tokens likely correspond to trivial questions that do not require deep reasoning. This filtering ensures we select prompts that generate appropriately complex reasoning traces. Second, we apply deduplication by removing duplicate instructions to prevent duplicate training samples. After filtering, we retain 3294 samples.

\subsection{Training Configuration}
\label{app:training}

All training are mainly implemented via the \texttt{verl} library~\citep{sheng2024hybridflow}, though we also have TRL~\citep{vonwerra2020trl} version. All experiments are conducted on a single node equipped with 8 NVIDIA H100 GPUs using Fully Sharded Data Parallel (FSDP2) for model parallelism.

The structure reward requires step embeddings computed from each rollout's \texttt{<think>} block. To avoid interfering with the main policy rollout (which runs under vLLM), we launch a standalone embedding server using \texttt{vllm.entrypoints.openai.api\_server} with the \texttt{Qwen/Qwen3-Embedding-0.6B} model on a dedicated port, consuming only 5\% of one GPU's memory. The reward function queries this server via HTTP for each generated trajectory.

Tables~\ref{tab:training_config_ppo} and~\ref{tab:training_config_grpo} summarize the training hyperparameters used in our PPO and GRPO experiments, respectively. Across both algorithms, the Math and Open-Ended settings mainly differ in the training data, maximum response length, total epochs, and clustering method.

\begin{table*}[ht]
    \centering
    \small
    \renewcommand{\arraystretch}{1.15}
    \begin{tabular}{lcc}
    \toprule
    \textbf{Hyperparameter} & \textbf{Math Setting} & \textbf{Open-Ended Setting} \\
    \midrule
    RL algorithm                      & PPO                    & PPO \\
    Advantage estimator               & GAE                    & GAE \\
    Training data                     & AIME 1983--2024        & OpenRubrics-v2 \\
    Base model                        & \texttt{Qwen/Qwen3-4B} & \texttt{Qwen/Qwen3-4B} \\
    Rollout engine                    & vLLM                   & vLLM \\
    Rollouts per prompt ($N$)         & 8                      & 8 \\
    Train batch size                  & 256                    & 256 \\
    Policy mini batch size            & 256                    & 256 \\
    Policy micro batch size / GPU     & 32                     & 32 \\
    Critic micro batch size / GPU     & 32                     & 32 \\
    Actor learning rate               & $1 \times 10^{-6}$     & $1 \times 10^{-6}$ \\
    Critic learning rate              & $1 \times 10^{-5}$     & $1 \times 10^{-5}$ \\
    Max prompt length                 & 1{,}024 tokens         & 1{,}024 tokens \\
    Max response length               & 8{,}192 tokens         & 4{,}096 tokens \\
    Filter overlong prompts           & True                   & True \\
    Truncation mode                   & \texttt{error}         & \texttt{error} \\
    Data shuffle                      & False                  & False \\
    KL loss coefficient               & 0.001                  & 0.001 \\
    KL loss type                      & \texttt{low\_var\_kl}  & \texttt{low\_var\_kl} \\
    Entropy coefficient               & 0                      & 0 \\
    Use KL in reward                  & False                  & False \\
    Critic warmup                     & 0                      & 0 \\
    Total epochs                      & 15                     & 5 \\
    Save frequency                    & 20                     & 20 \\
    Test frequency                    & 5                      & 5 \\
    Parallelism strategy              & FSDP2                  & FSDP2 \\
    Dynamic batch sizing              & True                   & True \\
    Gradient checkpointing            & True                   & True \\
    Activation offload                & True                   & True \\
    Rollout GPU memory utilization    & 0.7                    & 0.7 \\
    Embedding model                   & \texttt{Qwen/Qwen3-Embedding-0.6B} & \texttt{Qwen/Qwen3-Embedding-0.6B} \\
    Embedding server                  & standalone vLLM API server & standalone vLLM API server \\
    Clustering method                 & KMeans                 & HDBSCAN \\
    \bottomrule
    \end{tabular}
\caption{PPO training hyperparameters for the Math and Open-Ended settings.}
\label{tab:training_config_ppo}
\end{table*}

\begin{table*}[ht]
    \centering
    \small
    \renewcommand{\arraystretch}{1.15}
    \begin{tabular}{lcc}
    \toprule
    \textbf{Hyperparameter} & \textbf{Math Setting} & \textbf{Open-Ended Setting} \\
    \midrule
    RL algorithm                      & GRPO                   & GRPO \\
    Training data                     & AIME 1983--2024        & OpenRubrics-v2 \\
    Base model                        & \texttt{Qwen/Qwen3-4B} & \texttt{Qwen/Qwen3-4B} \\
    Rollout engine                    & vLLM                   & vLLM \\
    Rollouts per prompt ($N$)         & 8                      & 8 \\
    Train batch size                  & 256                    & 256 \\
    Policy mini batch size            & 256                    & 256 \\
    Policy micro batch size / GPU     & 32                     & 32 \\
    Actor learning rate               & $1 \times 10^{-6}$     & $1 \times 10^{-6}$ \\
    Max prompt length                 & 1{,}024 tokens         & 1{,}024 tokens \\
    Max response length               & 8{,}192 tokens         & 4{,}096 tokens \\
    Filter overlong prompts           & True                   & True \\
    Truncation mode                   & \texttt{error}         & \texttt{error} \\
    Data shuffle                      & False                  & False \\
    KL loss coefficient               & 0.001                  & 0.001 \\
    KL loss type                      & \texttt{low\_var\_kl}  & \texttt{low\_var\_kl} \\
    Entropy coefficient               & 0                      & 0 \\
    Use KL in reward                  & False                  & False \\
    Critic warmup                     & 0                      & 0 \\
    Total epochs                      & 15                     & 5 \\
    Save frequency                    & 20                     & 20 \\
    Test frequency                    & 5                      & 5 \\
    Parallelism strategy              & FSDP2                  & FSDP2 \\
    Dynamic batch sizing              & True                   & True \\
    Gradient checkpointing            & True                   & True \\
    Activation offload                & True                   & True \\
    Rollout GPU memory utilization    & 0.7                    & 0.7 \\
    Embedding model                   & \texttt{Qwen/Qwen3-Embedding-0.6B} & \texttt{Qwen/Qwen3-Embedding-0.6B} \\
    Embedding server                  & standalone vLLM API server & standalone vLLM API server \\
    Clustering method                 & KMeans                 & HDBSCAN \\
    \bottomrule
    \end{tabular}
\caption{GRPO training hyperparameters for the Math and Open-Ended settings.}
\label{tab:training_config_grpo}
\end{table*}

\section{Additional Ablations}
\label{app:additional-results}

\subsection{Full Response Length Comparison}
\label{app:length_analysis}

For completeness, we report the full response length comparisons against all available baselines for both the math and open ended settings.

\begin{table*}[ht]
\centering
\small
\setlength{\tabcolsep}{3pt}
\renewcommand{\arraystretch}{1.12}
\begin{tabular}{lcccccc}
\toprule
\textbf{Method} & \textbf{AIME25} & \textbf{AMC23} & \textbf{MATH500} & \textbf{Minerva} & \textbf{Avg} & \textbf{$\Delta$ vs. Base} \\
\midrule
Base & 7728.0 & 5884.1 & 4182.6 & 5151.7 & 5736.6 & 0.0 \\
PPO w/ Ground-Truth & 7047.7 & 4634.7 & 3124.9 & 4021.7 & 4707.3 & -1029.3 \\
\rowcolor{ourHi}
PPO w/ SARL & 7296.6 & 5069.4 & 3451.1 & 4485.4 & 5075.6 & -661.0 \\
GRPO w/ Ground-Truth & 6477.7 & 3956.6 & 2559.0 & 3143.8 & 4034.3 & -1702.3 \\
GRPO w/ EMPO & 6651.5 & 4058.6 & 2702.4 & 3327.9 & 4185.1 & -1551.5 \\
GRPO w/ TTRL & 6635.0 & 4057.8 & 2663.5 & 3258.5 & 4153.7 & -1582.9 \\
\rowcolor{ourHi}
GRPO w/ SARL & 6984.9 & 4506.9 & 3029.4 & 3829.9 & 4587.8 & -1148.8 \\
\bottomrule
\end{tabular}
\caption{Full average response length comparison on mathematical reasoning benchmarks.}
\label{tab:length_math_full}
\end{table*}

\begin{table*}[ht]
\centering
\small
\setlength{\tabcolsep}{4pt}
\renewcommand{\arraystretch}{1.12}
\begin{tabular}{lcc}
\toprule
\textbf{Method} & \textbf{Avg Tokens} & \textbf{$\Delta$ vs. Base} \\
\midrule
Base & 2677.91 & 0.00 \\
DPO & 2661.46 & -16.45 \\
EMPO & 2731.92 & +54.01 \\
\rowcolor{ourHi}
PPO w/ SARL & 2303.17 & -374.74 \\
\rowcolor{ourHi}
GRPO w/ SARL & 2285.87 & -392.04 \\
\bottomrule
\end{tabular}
\caption{Full average response length comparison on WildBench.}
\label{tab:length_openended_full}
\end{table*}

\subsection{Structure Reward Ablation Details}
\label{app:ablation_components}

We retrain Qwen3-4B under GRPO on the AIME math training set using each component of the structure reward in Eq.~\eqref{eq:reward} alone, while keeping all other settings identical. Both components individually yield substantial improvements over the base model. This observation is consistent with prior work that identifies clustering and path length as key structural properties associated with strong reasoning behavior~\citep{minegishi2025topology}.

\begin{table*}[ht]
\centering
\small
\setlength{\tabcolsep}{6pt}
\renewcommand{\arraystretch}{1.15}
\begin{tabular}{lccccc}
\toprule
\textbf{Reward variant} & \textbf{AIME25} & \textbf{AMC23} & \textbf{MATH500} & \textbf{Minerva} & \textbf{Avg ($\Delta$\%)} \\
\midrule
Base                                  & 31.67          & 82.81          & 90.10          & 53.45          & 64.51 \\
\midrule
SARL w/ $L$ only          & 40.42          & \textbf{87.81} & 92.92          & \textbf{62.68} & 70.96~(+10.0\%) \\
SARL w/ $C$ only          & \textbf{45.83} & 87.19          & 93.00          & 61.86          & 71.97~(+11.6\%) \\
\rowcolor{ourHi}
SARL w/ $C + L$           & \textbf{45.83} & 87.50          & \textbf{93.30} & 61.99          & \textbf{72.16~(+11.9\%)} \\
\bottomrule
\end{tabular}
\caption{Per-benchmark ablation of the two structure reward terms under GRPO on the math setting. $C$ denotes the average clustering coefficient and $L$ denotes the average shortest path length, with the reward components $C(\mathcal{G})/2$ and $1/(1+L(\mathcal{G}))$ defined in Eq.~\eqref{eq:reward}.}
\label{tab:ablation_components_full}
\end{table*}

\input{sections/further-analysis}

\section{Ethics and Broader Impact Statement}
\label{sec:ethics}
The paper does not involve human-subject data collection, personally identifiable information, or deployment in safety-critical settings. Potential risks include enabling stronger reasoning models that may also be misused in broader downstream applications. However, the paper primarily contributes to a label-free RL training technique rather than a new capability domain. We document datasets, implementation details, hyperparameters, and compute requirements, supporting external scrutiny while discouraging inappropriate use.

%% file: sections/further-analysis.tex
\section{Reward Robustness Analysis}
\label{app:analysis-results}
This section examines the structural reward, its combination with outcome supervision, and the robustness of the results across model and evaluation choices. Unless specified otherwise, math columns are accuracy percentages and ``Avg'' is the mean over AIME25, AMC23, MATH500, and Minerva. Relative changes in parentheses are measured against the corresponding base model. The training comparisons use single runs; we do not attach statistical significance claims to small differences.

\subsection{Structural Reward Diagnostics}
\label{app:analysis-reward-hacking}
\paragraph{Best-of-$n$ proxy stress test.}
For each of the 30 AIME25 problems, we sampled 256 responses from the base Qwen3-4B policy and 256 from the SARL-trained policy (15{,}360 responses in total). For each candidate pool, we selected the response with the highest structure reward and checked its answer using an independent ground-truth verifier. We averaged over 1{,}000 random permutations per problem. Random selection and oracle pass@$n$ provide reference points. Table~\ref{tab:analysis-bon} shows that selected structure reward rises with pool size while verifier accuracy increases substantially through $n=256$, with a small fluctuation between $n=64$ and $n=128$ for the trained policy. There is no sustained accuracy collapse over this tested range. The oracle gap for the trained policy remains substantial, so this test does not show that the proxy identifies every correct response.

\par\begingroup\centering
\centering\small\setlength{\tabcolsep}{5pt}
\begin{tabular}{llrrrr}
\toprule
Policy & $n$ & Random acc. & SR-selected acc. & Oracle pass@$n$ & Selected SR \\
\midrule
Base & 1 & 22.97 & 22.97 & 22.97 & 0.165 \\
 & 8 & 22.64 & 36.98 & 38.51 & 0.298 \\
 & 32 & 22.53 & 44.47 & 45.74 & 0.379 \\
 & 64 & 22.94 & 47.39 & 47.86 & 0.417 \\
 & 128 & 22.56 & 49.19 & 49.22 & 0.449 \\
 & 256 & 22.89 & 50.00 & 50.00 & 0.473 \\
\midrule
\rowcolor{ourHi}
SARL & 1 & 41.90 & 41.90 & 41.90 & 0.354 \\
\rowcolor{ourHi}
 & 8 & 41.99 & 55.56 & 61.00 & 0.575 \\
\rowcolor{ourHi}
 & 32 & 41.94 & 59.40 & 69.69 & 0.713 \\
\rowcolor{ourHi}
 & 64 & 41.99 & 59.97 & 72.32 & 0.767 \\
\rowcolor{ourHi}
 & 128 & 41.88 & 59.77 & 74.92 & 0.802 \\
\rowcolor{ourHi}
 & 256 & 41.97 & 60.00 & 80.00 & 0.817 \\
\bottomrule
\end{tabular}
\captionof{table}{Best-of-$n$ selection on AIME25. Accuracy and oracle pass@$n$ are percentages; SR is structure reward.}
\label{tab:analysis-bon}
\par\endgroup

\paragraph{Graph statistics.}
Table~\ref{tab:analysis-graph} compares all 7{,}680 responses per policy from the same AIME25 samples. Mean density increases after training but remains well below one, and complete graphs occur in 0.143\% of trained-policy responses. These observations address collapse toward fully connected graphs in this setting, but do not by themselves establish semantic faithfulness.

\par\begingroup\centering
\centering\small
\begin{tabular}{lr>{\columncolor{ourHi}}r}
\toprule
Metric & Base & SARL \\
\midrule
Mean structure reward & 0.164 & 0.354 \\
Mean graph density & 0.129 & 0.278 \\
Full-clique rate (\%) & 0.000 & 0.143 \\
Normalized clique number & 0.110 & 0.243 \\
Transitivity & 0.135 & 0.288 \\
State-revisit rate & 0.200 & 0.438 \\
\bottomrule
\end{tabular}
\captionof{table}{Topology of sampled AIME25 responses before and after SARL training.}
\label{tab:analysis-graph}
\par\endgroup

\subsection{Combining Structural and Outcome Rewards}
\label{app:analysis-combined}
We also trained with an equally weighted sum of ground-truth accuracy reward and structure reward on math. Table~\ref{tab:analysis-combined} compares the combined objective with either reward alone. The combination has the highest PPO average (71.06), whereas SARL alone has the highest GRPO average (72.16). The differences are small relative to the gains over Base and lack multi-seed uncertainty estimates. This experiment applies to verifiable math only; open-ended prompts have no ground-truth accuracy reward.

\par\begingroup\centering
\centering\small\setlength{\tabcolsep}{5pt}
\begin{tabular}{lrrrrr}
\toprule
Method & AIME25 & AMC23 & MATH500 & Minerva & Avg (relative $\Delta$) \\
\midrule
Base & 31.67 & 82.81 & 90.10 & 53.45 & 64.51 \\
PPO w/ Ground-Truth & 41.67 & 86.56 & 92.75 & 59.74 & 70.18 (+8.8\%) \\
\rowcolor{ourHi}
PPO w/ SARL & 42.92 & 85.00 & 92.53 & 61.08 & 70.38 (+9.1\%) \\
\rowcolor{ourHi}
PPO w/ Ground-Truth + SARL & 42.92 & 86.39 & 93.60 & 61.31 & 71.06 (+10.1\%) \\
GRPO w/ Ground-Truth & 46.67 & 84.38 & 93.15 & 62.45 & 71.66 (+11.1\%) \\
\rowcolor{ourHi}
GRPO w/ SARL & 45.83 & 87.50 & 93.30 & 61.99 & 72.16 (+11.9\%) \\
\rowcolor{ourHi}
GRPO w/ Ground-Truth + SARL & 43.75 & 87.50 & 93.20 & 61.86 & 71.58 (+11.0\%) \\
\bottomrule
\end{tabular}
\captionof{table}{Math accuracy under separate and combined rewards. The combined reward is $0.5$ ground-truth accuracy reward $+\,0.5$ structure reward.}
\label{tab:analysis-combined}
\par\endgroup

\subsection{Sensitivity to Representation and Model Choice}
\label{app:analysis-robustness}
Table~\ref{tab:analysis-clustering} reports GRPO math results with HDBSCAN in place of KMeans, and with \texttt{microsoft/harrier-oss-v1-0.6b} in place of \texttt{Qwen/Qwen3-Embedding-0.6B}, changing one component at a time. Both alternatives retain substantial gains over Base. In the reported training run, GRPO with SARL took 11 h 27 m 28 s versus 11 h 0 m 26 s for ground-truth reward training on the stated hardware (Appendix~\ref{app:training}).

\par\begingroup\centering
\centering\small\setlength{\tabcolsep}{5pt}
\begin{tabular}{llrrrrr}
\toprule
Variant & Setting & AIME25 & AMC23 & MATH500 & Minerva & Avg (relative $\Delta$) \\
\midrule
Base & -- & 31.67 & 82.81 & 90.10 & 53.45 & 64.51 \\
\rowcolor{ourHi}
Clustering & HDBSCAN & 43.33 & 86.88 & 92.53 & 61.67 & 71.10 (+10.2\%) \\
\rowcolor{ourHi}
Clustering & KMeans & 45.83 & 87.50 & 93.30 & 61.99 & 72.16 (+11.9\%) \\
\rowcolor{ourHi}
Embedding & Qwen3-Embedding-0.6B & 45.83 & 87.50 & 93.30 & 61.99 & 72.16 (+11.9\%) \\
\rowcolor{ourHi}
Embedding & harrier-oss-v1-0.6b & 43.75 & 87.50 & 92.71 & 61.86 & 71.46 (+10.8\%) \\
\bottomrule
\end{tabular}
\captionof{table}{Clustering and embedding ablations for Qwen3-4B under GRPO on math. The KMeans and Qwen embedding rows are the same reference run.}
\label{tab:analysis-clustering}
\par\endgroup

We further evaluated Qwen3-8B and Llama-3.2-3B-Instruct. For the non-thinking Llama model, we used reasoning-unit segmentation of its generated text, preserving equations, lists, steps, and cases, rather than requiring an explicit \texttt{<think>} block. Tables~\ref{tab:analysis-scale-math} and~\ref{tab:analysis-scale-wild} report the corresponding math and WildBench results. Gains transfer to these tested models, although the Llama math gain from SARL is smaller than its ground-truth reward gain. The WildBench score is a weighted macro score over five categories; relative changes use the base score for each model family.

\par\begingroup\centering
\centering\small\setlength{\tabcolsep}{5pt}
\resizebox{\textwidth}{!}{%
\begin{tabular}{llrrrrr}
\toprule
Model & Method & AIME25 & AMC23 & MATH500 & Minerva & Avg (relative $\Delta$) \\
\midrule
Qwen3-8B & Base & 31.67 & 78.13 & 87.93 & 54.60 & 63.08 \\
 & GRPO w/ Ground-Truth & 41.67 & 86.25 & 93.13 & 63.92 & 71.24 (+12.9\%) \\
\rowcolor{ourHi}
 & GRPO w/ SARL & 40.00 & 85.63 & 92.75 & 62.55 & 70.23 (+11.3\%) \\
\midrule
Llama-3.2-3B-Instruct & Base & 0.00 & 17.50 & 39.43 & 17.88 & 18.70 \\
 & GRPO w/ Ground-Truth & 0.42 & 23.44 & 45.10 & 24.40 & 23.34 (+24.8\%) \\
\rowcolor{ourHi}
 & GRPO w/ SARL & 1.67 & 22.50 & 39.75 & 18.66 & 20.65 (+10.4\%) \\
\bottomrule
\end{tabular}
}
\captionof{table}{Math results across model scales and families.}
\label{tab:analysis-scale-math}
\par\endgroup

\par\begingroup\centering
\centering\small\setlength{\tabcolsep}{4pt}
\resizebox{\textwidth}{!}{%
\begin{tabular}{llrrrrrrr}
\toprule
Model & Method & Creative & Planning & Math & Info & Coding & WB Score (relative $\Delta$) & Length \\
\midrule
Qwen3-8B & Base & 55.87 & 43.05 & 24.84 & 53.51 & 24.72 & 37.39 & 11170.99 \\
 & DPO & 55.13 & 41.97 & 23.17 & 52.51 & 25.47 & 36.72 (-1.8\%) & 11084.02 \\
\rowcolor{ourHi}
 & GRPO w/ SARL & 60.73 & 50.70 & 33.81 & 59.01 & 36.70 & 45.83 (+22.6\%) & 9419.47 \\
\midrule
Llama-3.2-3B & Base & 39.69 & 28.01 & 10.71 & 33.17 & 11.00 & 21.93 & 3802.48 \\
 & DPO & -5.65 & -14.25 & -32.46 & 0.62 & -30.46 & -19.10 (-187.1\%) & 3008.39 \\
\rowcolor{ourHi}
 & GRPO w/ SARL & 40.26 & 29.61 & 15.40 & 34.60 & 13.08 & 24.18 (+10.3\%) & 2276.47 \\
\bottomrule
\end{tabular}
}
\captionof{table}{WildBench results across model scales and families. Negative WB Scores are possible because the score is centered on the judge's mid-level rating. Length is average tokens.}
\label{tab:analysis-scale-wild}
\par\endgroup

\subsection{Open-Ended Evaluation Robustness}
\label{app:analysis-evaluation}
We evaluated Qwen3-4B on AlpacaEval as a second open-ended benchmark (Table~\ref{tab:analysis-alpaca}). We also re-scored WildBench outputs using three judges (Table~\ref{tab:analysis-judges}). We report the standard errors accompanying the AlpacaEval metrics. WildBench scores are not directly comparable across judges, so the second table gives ranks and absolute changes from Base within each judge. Both SARL variants rank first or second under each judge, but the size of their gains depends on the evaluator. These automated evaluations do not replace human assessment.

\par\begingroup\centering
\centering\small\setlength{\tabcolsep}{6pt}
\begin{tabular}{lrrrrr}
\toprule
Model & Win rate (\%) & SE (win) & LC win rate (\%) & SE (LC) & Avg length \\
\midrule
Base & 38.39 & 1.72 & 24.04 & 0.08 & 6733 \\
DPO & 38.81 & 1.72 & 24.27 & 0.10 & 6577 \\
\rowcolor{ourHi}
PPO w/ SARL & 46.26 & 1.76 & 30.37 & 0.12 & 5332 \\
GRPO w/ EMPO & 39.75 & 1.73 & 27.56 & 0.12 & 7186 \\
\rowcolor{ourHi}
GRPO w/ SARL & 43.41 & 1.75 & 26.97 & 0.17 & 5187 \\
\bottomrule
\end{tabular}
\captionof{table}{Qwen3-4B on AlpacaEval. LC denotes length-controlled win rate; SE denotes reported standard error.}
\label{tab:analysis-alpaca}
\par\endgroup

\par\begingroup\centering
\centering\small\setlength{\tabcolsep}{5pt}
\begin{tabular}{lrrrrrr}
\toprule
 & \multicolumn{3}{c}{Rank} & \multicolumn{3}{c}{WB Score $\Delta$ vs. Base} \\
\cmidrule(lr){2-4}\cmidrule(lr){5-7}
Method & GPT-4o & GPT-5.6-sol & Claude-5-sonnet & GPT-4o & GPT-5.6-sol & Claude-5-sonnet \\
\midrule
Base & 4 & 3 & 3 & 0.00 & 0.00 & 0.00 \\
DPO & 3 & 5 & 5 & +0.43 & -0.21 & -7.98 \\
GRPO w/ EMPO & 5 & 4 & 4 & -0.71 & -0.05 & -7.07 \\
\rowcolor{ourHi}
GRPO w/ SARL & 2 & 1 & 2 & +9.10 & +5.34 & +1.82 \\
\rowcolor{ourHi}
PPO w/ SARL & 1 & 2 & 1 & +10.35 & +4.93 & +4.36 \\
\bottomrule
\end{tabular}
\captionof{table}{Qwen3-4B WildBench sensitivity to the judge. Rank 1 is best. DPO uses preference labels; $\Delta$ is an absolute, within-judge difference from Base.}
\label{tab:analysis-judges}
\par\endgroup